# Data Augmentation to Address Out-of-Vocabulary Problem in Low-Resource Sinhala-English Neural Machine Translation

**Aloka Fernando**
University of Moratuwa
Sri Lanka
alokaf@cse.mrt.ac.lk

**Surangika Ranathunga**
University of Moratuwa
Sri Lanka
surangika@cse.mrt.ac.lk

## Abstract

Out-of-Vocabulary (OOV) is a problem for Neural Machine Translation (NMT). OOV refers to words with a low occurrence in the training data, or to those that are absent from the training data. To alleviate this, word or phrase-based Data Augmentation (DA) techniques have been used. However, existing DA techniques have addressed only one of these OOV types and limit to considering either syntactic constraints or semantic constraints. We present a word and phrase replacement-based DA technique that consider both types of OOV, by augmenting (1) rare words in the existing parallel corpus, and (2) new words from a bilingual dictionary. During augmentation, we consider both syntactic and semantic properties of the words to guarantee fluency in the synthetic sentences. This technique was experimented with low resource Sinhala-English language pair. We observe with only semantic constraints in the DA, the results are comparable with the scores obtained considering syntactic constraints, and is favourable for low-resourced languages that lacks linguistic tool support. Additionally, results can be further improved by considering both syntactic and semantic constraints.

## 1 Introduction

With the impressive results they produced, NMT systems have become the state-of-the-art solution for the problem of Machine Translation (MT). Although MT task is an open-vocabulary problem, the NMT solutions are limited to a fixed vocabulary, constrained by the size of the parallel corpus (Luong and Manning, 2016). This limited vocabulary gives rise to the OOV problem, which is two-fold. Firstly, there can be rare words, with significantly low occurrence in the training data (Sennrich et al., 2016). Secondly, there are words that are totally absent in the training corpus (Peng et al., 2020). In both these scenarios, vanilla NMT models fail to produce reliable translation outputs (Bahdanau et al., 2015).

In this paper, we address the problem of OOV with respect to the Sinhala-English (Si-En) language pair. Sinhala is a morphologically rich, low-resource Indo-Aryan language. According to language classification by Joshi et al. (2020), Sinhala fits into class 0, because it has exceptionally limited resources and labeled data. For such a low resource language pair, the OOV problem is severe. DA in the context of MT is to induce a synthetic parallel corpus, by exploiting monolingual data or bilingual lexicons. Recently, the technique had been explored to address the OOV problem. Fadaee et al. (2017) proposed a word replacement-based method and they generated the synthetic sentences by replacing a selected word in the existing training sentence with a rare word. Tennage et al. (2018b) extended this by incorporating linguistic features Part-of-Speech (POS) and morphology to validate the replacement syntactically. However, the synthetic sentences may not necessarily be semantically correct. Peng et al. (2020) generated an in-domain pseudo-parallel corpus, by replacing terms from an in-domain bilingual dictionary into identified words in the parallel sentences. This technique considered only semantic information and preserved the semantic correctness in the replacement. However, they disregarded measures to validate the replacement syntactically.

In our research, we generate a pseudo parallel corpus with synthetic sentences that augments rare words in the training corpus, as well as new terms from a bilingual dictionary, by replacing them in suitable contexts in the existing parallel sentences. In contrast to the previous research, we consider both syntactic and semantic constraints in the DA technique. As syntactic constraints, we use POS and morphological information similar to Tennage et al. (2018b) and word embedding-based semantic constraints similar to Peng et al. (2020).

Firstly, we empirically show that the scores obtained by only using semantic constraints are comparable to the scores obtained by using syntactic constraints. This is a very useful observation, as it shows the potential of embeddings. This suggests that the low resource languages that lack linguistic resources can obtain the same benefit from this DA technique by relying on semantic information only. As monolingual data are available for low resource languages word embeddings can be generated easily for such languages. Secondly, we show that by combining both syntactic and semantic constraints, the results can be improved further.

## 2 Related Work

### 2.1 Data Augmentation to Address the OOV Problem

Data augmentation research in the context of MT can be categorized as techniques using (1) word or phrase-based augmentation (2) back-translation and (3) parallel corpus mining (Ranathunga et al., 2021). In this section, we consider only word or phrase-based augmentation.

Fadaee et al (2017) are the first to explicitly focus on a word replacement method to augment the rare words to address the OOV problem. In this technique, for a considered parallel sentence pair, a common word in the source sentence is replaced by a rare word in the source language. The synthetic target side sentence is obtained by replacing the aligned target side common word, with the translation of the source rare word. As the synthetically generated sentences lacked fluency, Tennage et al. (2018b) incorporated linguistic constraints to validate the replacement. Here the word was replaced if the rare word and the common word identified in the sentence agreed in terms of POS and morphology. In a similar work, Duan et al. (2020) relied on dependency information to determine the suitable word to be replaced in the sentence. In both these techniques, although the syntactic correctness of the synthetic sentence was preserved, semantic correctness could not be guaranteed. Additionally, the sub-optimal nature of these linguistic tools can further add noise to the process.

The DA techniques that use dictionaries expand the vocabulary of the NMT models and are in favour of addressing the OOV problem as well. Nag et al. (2020) used a bilingual dictionary to translate target-side monolingual sentences word-to-word, in order to obtain the source side synthetic sentences. However, for a morphologically rich language, such a method would lead to sub-optimal results, as the dictionary terms are mostly in the base form and do not provide translations for the inflected terms. Alternatively, Peng et al (2020) used an in-domain dictionary to induce synthetic parallel sentences from an out-domain parallel corpus using a phrase-replacement augmentation, considering only semantic similarity measurements. First, they filtered the top most similar sentences, considering the semantic similarity between the source side dictionary term and the source side sentences from the out-domain parallel corpus. Then they determined the noun phrase to be replaced in the candidate sentence, based on the semantic similarity between the source dictionary term and the noun phrases in the sentence. However, the task remains sub-optimal as the technique does not guarantee a syntactically correct replacement.

### 2.2 Sinhala related NMT

For Sinhala language, linguistic processing tools and resources such as morphological analyzers, dependency parsers, or annotated datasets are scarce (de Silva, 2019). As a result, Sinhala-related NMT research still lags in achieving state-of-the-art results.

Initial research on Sinhala-related NMT was between the Sinhala-Tamil language pair (Tennage et al., 2017) which was subsequently improved with transliteration and Byte-pair-Encoding (BPE) by Tennage et al. (2018a). In the recent work, the transformer architecture with BPE has shown significant gains for the Sinhala-Tamil pair (Pramodya et al., 2020), and for the English-Sinhala pair (Fon-seka et al., 2020). Sinhala related NMT obtained improvements with back-translation related techniques (Nissanka et al., 2020; Pushpananda, 2019).

The Sinhala-English FLORES dataset (Guzmán et al., 2019) and OPUS-100 (Zhang et al., 2020) dataset were released with the objective of supporting low resource NMT and multilingual NMT, respectively. Although the datasets were used in recent research work (Wang et al., 2020; Li et al., 2020), they treated the languages as a blank-box and did not incorporate language dependant constraints. Further, those solutions did not explicitly address the OOV problem.

## 3 Methodology

Our DA solution follows a word-replacement based augmentation strategy, which incorporates the potential of semantic information, as well as syntactic information in producing a pseudo parallel corpus. This is in contrast to considering only one of these constraints as done in previous research. This augmentation is two-fold:

- Rare word augmentation - For a selected sentence pair in the existing training corpus, a candidate word in the source side sentence is replaced by a rare word identified from the source-side of the parallel corpus, confirming both syntactic and semantic constraints. This produces the source-side synthetic sentence. Similarly, the target side synthetic sentence is obtained by replacing the aligned word or phrase from the target sentence with the translation of the rare word.

- Dictionary augmentation - To obtain a synthetic source side sentence from a considered parallel sentence pair, a selected word from the source side parallel sentence is replaced by a source side dictionary term confirming the same syntactic and semantic constraints. The target side synthetic sentence is obtained by replacing the corresponding aligned word or phrase, with the target side dictionary term.

### 3.1 Rare Word Augmentation

In this data augmentation technique, rare words are substituted in existing parallel sentences to provide novel contexts. The rare word augmentation process is shown in Figure 1, and the step-wise process is described in the following sections.

#### 3.1.1 Obtain Rare words:

We identify the rare words from the source side of the parallel corpus as done by Fadaee et al. (2017) and Tennage et al.(2017). The words with an occurrence less than a threshold ($T_R$) are considered as rare words. A word-alignment model is trained on the parallel corpus to obtain the corresponding target side rare word. In Figure 1 this is the Step 1.

#### 3.1.2 Word/Sentence Embeddings:

Pre-trained word embeddings are rich in terms of capturing both language constraints as well as word-related constraints. This has been empirically proven for Sinhala as well (2020). However, for Sinhala, the potential of word embeddings has not been explored in the context of data augmentation. Therefore in our work, we incorporate semantic information in two instances: (1) to filter out candidate sentences from the existing parallel corpus to conduct the rare word replacement as described in section 3.1.3, and (2) to identify the word to be replaced in this sentence as described in section 3.1.4.

Although the embeddings are widely used in Natural Language Processing (NLP) tasks, different embedding types capture the word constraints differently. This is further evident when obtaining the most similar words (Mikolov et al., 2013). For example, some embeddings are better suited to capture the syntactic similarity between words (e.g. run - running) whereas others are better suited to capture semantic similarity (e.g. sing - chant).

Artetxe et al.(2018) empirically proved that, by conducting a linear transformation on the embedding vectors as a post-processing step, they can be used to capture the intended type of similarity (semantic similarity or syntactic similarity) between words. Therefore to optimize the embeddings for the DA task, we post-process them according to Artetxe et al.(2018) using a parameter value (alpha). As the value needs to be determined experimentally, the word embeddings were post-processed using different alpha values and were used in DA experiments. The alpha value which returned the best score was selected to post-process the word embeddings in the subsequent experiments.

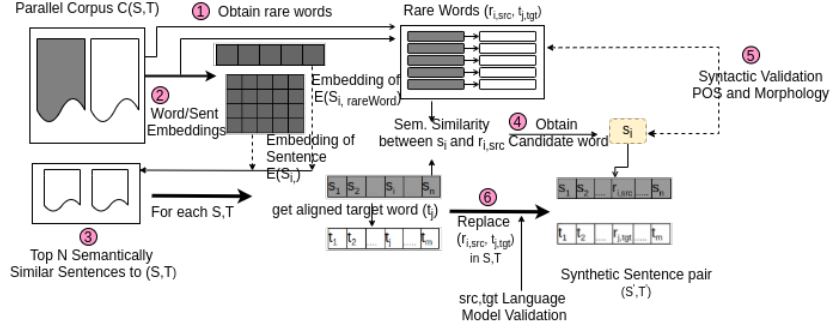

Figure 1: Data Augmentation Process

Sentence embeddings were obtained by averaging the post-processed embeddings of individual words of the sentence. This is step 2 in Figure 1.

### 3.1.3 Obtaining Candidate sentences

It is important to find out the most suitable sentences to replace the rare word, from the rest of the sentences. First we consider the semantic similarity between the source side training sentences and the sentence containing the source side rare word. Then we select the topmost similar sentences as candidate sentences for the DA. This is step 3 in Figure 1.

### 3.1.4 Obtaining the Candidate Word

In step 4 in Figure 1, to identify the word to be replaced, we calculate the cosine similarity between the source side rare word and each word in the source sentence. The word with the maximum cosine similarity is considered as the candidate word for the replacement.

### 3.1.5 Syntactic Validation

In step 5, the identified candidate word is further checked for syntactic agreement with the rare word. Similar to Tennage et al. (2018b) the POS and morphology agreement is checked for Sinhala words. For English words, we consider only number agreement. Further syntactic constraints such as dependency rules (Duan et al., 2020) have not been considered, since Sinhala does not have a syntactic parser.

### 3.1.6 Generating Synthetic Sentences

In step 6, the candidate word identified in step 5 is replaced with the source rare word to produce the source side synthetic sentence.

A word-alignment model is used to identify the target side word or phrase to be replaced in the target side sentence, and this is replaced by the target side rare word to obtain the synthetic target side sentence.

Then the replacement context is further validated by a Language Model (LM), trained on monolingual corpora in the respective languages. The source-side tri-gram context is scored using the source side LM and the target side tri-gram context is scored by the target side LM respectively.

The replacement is accepted if the ratio between the LM score in the synthetic sentence and LM score in the original sentence, exceeds a threshold for both source and target sides.

### 3.2 Dictionary Augmentation

The dictionary augmentation algorithm is similar to rare word augmentation, except for step 3 in Figure 1. According to DA results in Table 4, since sentence filtration did not give an improvement, this step was omitted in the dictionary augmentation.

As a result, all the source-side sentences were considered as candidate sentences for the dictionary term replacement. Here too, the obtained embeddings were post-processed prior to considering for the DA experiments. From step 4 onwards, the dictionary augmentation process is identical to the rare word augmentation process as illustrated in Figure 1.

### 3.3 Combined Solution

Finally, as a combined solution, we merge the DA sets, augmenting rare words and dictionary terms (as new vocabulary), and evaluate its effectiveness.

# 4 Experiments

## 4.1 Dataset

We used a Sinhala-English parallel corpus specific to the government document domain, which is an improved dataset of Fernando et al. (2020). The corpus statistics are given in Table 1.

|  | Train | Valid |
|---|---|---|
| No. Sentences | 54914 | 1623 |
| No. Words(En) | 553002 | 23578 |
| No. Words(Si) | 535185 | 22721 |

Table 1: Parallel Corpus Statistics of Training and Validation sets

The testset statistics are given in Table 2. We have selected different test sets, containing different numbers of rare words and dictionary terms to analyse the effectiveness of the technique, under different OOV conditions. Here the Dic.Terms(OOV) refer to the number of dictionary words in the testsets that are not a part of the training data.

|  | TS1 | TS2 | TS3 |
|---|---|---|---|
| No. Sentences | 1603 | 1462 | 1438 |
| **Sinhala** | | | |
| No. Words | 18513 | 28918 | 26308 |
| Unique Words | 4520 | 5341 | 5057 |
| Rare Words | 76 | 133 | 127 |
| Dic.Terms | 502 | 596 | 594 |
| Dic.Terms(OOV) | 11 | 17 | 23 |
| **English** | | | |
| No. Words | 19248 | 30437 | 27815 |
| Unique Words | 4237 | 4956 | 4865 |
| Rare words | 55 | 55 | 68 |
| Dic.Terms | 1314 | 1804 | 1739 |
| Dic.Terms(OOV) | 58 | 108 | 99 |

Table 2: Testset Statistics

The statistics of the Sinhala and English monolingual data (Isuranga et al., 2020) used to train the language models in the respective languages are detailed in Table 3. The monolingual data includes publicly available government documents, common crawl data and news data. We used SRILM (Stocke, 2011) toolkit to generate the language models.

|  | English | Sinhala |
|---|---|---|
| No of Sentences | 1,286,945 | 1,163,675 |
| No of Words | 51,193,388 | 48,283,636 |

Table 3: Monolingual Corpus Statistics

For the dictionary augmentation, we used an English-Sinhala dictionary [1] extracted from public data with a total of 23660 terms.

## 4.2 NMT Experimental Setup

We used the encoder-decoder with attention NMT architecture (Bahdanau et al., 2015) for our experiments. However, the proposed DA technique is independent of the NMT architecture.

We used the OpenNMT (Klein et al., 2018) toolkit(PyTorch version) for the NMT experiments. The experiments were conducted on Google Colaboratory (Colab) on an Nvidia K80 GPU with 8GB RAM. The NMT encoder is a 2-layer bidirectional Long Short Term Memory (LSTM) unit and the decoder is a 2-layer LSTM, with global attention. We used a batch size of 32, a drop-out probability of 0.4, and Adam Optimizer during training. We used the Moses tokenizer (Koehn et al., 2007) and a custom-built tokenizer (Farhath et al., 2018) to tokenize sentences for English and Sinhala languages respectively. We report the single-model tokenized BLEU scores on the testsets, using multi-bleu.perl (Papineni et al., 2002) script. To minimize any effect of fluctuations, each NMT experiment was executed three times and the average score is reported.

## 4.3 Baseline Models

Our baseline model was trained with 54K parallel sentences without DA. ie. Baseline[train54K].

To benchmark, we recreated Fadaee et al. (2017) and Peng et al. (2020) baselines using our dataset. For comparison with baseline, NMT models were trained by replacing rare words and dictionary terms *randomly*, in existing sentences without considering any syntactic or semantic constraints. And secondly, augmenting with, random samples of 10K, 25K, 35K from training. Thereby we analyze whether mere duplication improves the results. The results for the above experiments are shown in Table 4.

---

[1]https://www.maduraonline.com/

| Experiment | Aug. Sent. | Si→ En(BLEU) | | | Aug. Sent. | En → Si (BLEU) | | |
|---|---|---|---|---|---|---|---|---|
| | | TS1 | TS2 | TS3 | | TS1 | TS2 | TS3 |
| Baseline[train54K] | | 22.47 | 21.22 | 26.82 | | 20.61 | 19.33 | 24.97 |
| Baseline(Fadaee et al., 2017) | 10947 | 22.76 | 21.28 | 26.89 | 13675 | 20.80 | 18.95 | 24.62 |
| Baseline(Peng et al., 2020) | 12447 | 22.63 | 21.06 | 26.62 | 13675 | 20.49 | 19.30 | 25.37 |
| **Random Duplicating** | | | | | | | | |
| Baseline+randDuplicate10K | 10000 | 22.40 | 20.89 | 26.30 | 10000 | 20.39 | 19.12 | 24.48 |
| Baseline+randDuplicate25K | 25000 | 22.65 | 21.29 | 27.05 | 25000 | 21.00 | 19.44 | 25.38 |
| Baseline+randDuplicate35K | 35000 | 22.59 | 21.05 | 26.76 | 35000 | 20.25 | 19.38 | 25.33 |
| **Random Replacement** | | | | | | | | |
| Baseline+randRareWords10K | 10000 | 22.26 | 20.53 | 26.25 | 10000 | 20.67 | 19.33 | 25.11 |
| Baseline+randDictionary10K | 10000 | 22.50 | 20.77 | 26.56 | 10000 | 20.61 | 18.60 | 24.60 |
| **Linguistic constraints** | | | | | | | | |
| Baseline+pos | 2276 | 22.56 | 21.44 | 27.46 | 2587 | 20.76 | 19.44 | 25.33 |
| Baseline+pos+morph | 1560 | 22.40 | 21.50 | 27.43 | 2760 | 20.99 | 19.33 | 25.35 |
| **Word Similarity** | | | | | | | | |
| Baseline+wordSim$_{wo\ pp}$ | 8684 | 22.18 | 21.23 | 26.65 | 7792 | 20.48 | 18.78 | 25.08 |
| Baseline+wordSim | 7667 | 22.35 | 21.39 | 27.28 | 7544 | 21.08 | 19.23 | 25.12 |
| Baseline+wordSim+pos | 1789 | **22.88** | **21.84** | **27.73** | 3780 | 20.88 | **19.51** | 25.56 |
| Baseline+wordSim+pos+morph | 927 | 22.34 | 21.47 | 27.55 | 1780 | 20.89 | 19.47 | 25.53 |
| **Word Similarity+ sent Similarity** | | | | | | | | |
| Baseline+wordSim+sentSim | 7518 | 22.57 | 21.40 | 27.11 | 6642 | 20.97 | 19.07 | 25.13 |
| Baseline+wordSim +sentSim+pos+morph | 854 | 22.42 | 21.56 | 27.64 | 130 | **21.18** | 19.40 | **25.71** |

Table 4: Rare word Augmentation Results. *Baseline[train54K]*:Baseline NMT considered for our experiments.

### 4.4 Augmentation of Rare Words

Following Tennage (2017), words with frequency threshold $T_R = 1$ in the training corpus were considered as rare words. A total of 3133 and 2370 valid rare words were identified from the Sinhala and English sides respectively. GIZA++ automatic word alignment algorithm (Och and Ney, 2003) was used to determine the corresponding word or phrase from the target side.

To obtain POS and morphology information for Sinhala, the TnT POS Tagger (Fernando and Ranathunga, 2018) and sin-morphy[2] were used respectively. For the English side, the python NLP library Spacy[3] was used. The tri-gram LM threshold was chosen as 0.6 for context validation.

We used fastText embeddings (Bojanowski et al., 2016) to obtain word embeddings for Sinhala and English words. However, the fastText embeddings return syntactically similar words, rather than semantically similar words. As we need semantically similar words for our DA task, the fastText embeddings were post-processed according to Artetxe (2018).

To determine the suitable alpha value for post-processing, we conducted the DA experiment by considering only word similarity, with the post-processed embeddings using different alpha values. Experimentally the alpha values were selected as -0.15 and 0.15 for Sinhala and English, respectively. In Table 4, the rows Baseline+wordSim$_{w/o\ pp}$ and Baseline+wordSim show that post-processing the embeddings with these identified alpha values, improve the BLEU score with a maximum of +0.63 and +0.60 in the respective translation directions.

An ablation study was conducted by incorporating (1) syntactic constraints only, (2) semantic constraints only, and then by (3) combining both syn-tactic and semantic features in the DA experiments. Our objective was to identify the most suitable feature combination for the augmentation task. The results obtained are shown in Table 4.

---
[2]http://nlp-tools.uom.lk/sin-morphy/
[3]https://spacy.io/

| Experiment | Aug. Sent. | Si → En(BLEU) | | | Aug. Sent. | En → Si (BLEU) | | |
| --- | --- | --- | --- | --- | --- | --- | --- | --- |
| | | TS1 | TS2 | TS3 | | TS1 | TS2 | TS3 |
| Baseline[train54K] | | 22.47 | 21.22 | 26.82 | | 20.61 | 19.33 | 24.97 |
| Baseline(Fadaee et al., 2017) | 35901 | 21.59 | 19.36 | 22.70 | 49211 | 20.31 | 17.59 | 22.39 |
| Baseline(Peng et al., 2020) | 4856 | 22.28 | 20.76 | 26.17 | 5709 | 20.85 | 19.24 | 24.75 |
| **Linguistic constraints** | | | | | | | | |
| Baseline+pos | 26940 | 22.37 | 20.84 | 25.49 | 15201 | 20.63 | 18.41 | 24.15 |
| Baseline+pos+morph | 18770 | **22.65** | 21.25 | 26.38 | 15201 | 20.50 | 18.76 | 24.26 |
| **Word Similarity** | | | | | | | | |
| Baseline+wordSim | 32170 | 21.57 | 20.39 | 24.96 | 57288 | 19.95 | 18.20 | 22.04 |
| Baseline+wordSim+pos | 18209 | 21.51 | **21.29** | **26.40** | 25651 | 20.26 | 18.52 | 23.64 |
| Baseline+wordSim+pos+morph | 12594 | 22.07 | 20.87 | 26.21 | 6721 | **21.02** | **19.42** | **25.68** |
| **Combined Experiment** | | | | | | | | |
| Baseline+rareWord+dicTerm | 19998 | 22.17 | 20.69 | 26.13 | 7031 | **20.66** | 19.31 | **25.55** |

Table 5: Dictionary Augmentation Results. *Baseline[train54K]*:Baseline NMT considered for our experiments.

### 4.5 Augmentation of Dictionary

The terms in the Sinhala side of the dictionary were mostly phrases. For those terms, the embeddings were obtained by averaging the individual fastText word embeddings. The dictionary term embeddings were also post-processed using the same alpha values obtained earlier.

The experiments for dictionary augmentation were conducted in the same manner by considering syntactic constraints only, semantic constraints only and by combining both types of constraints respectively. The results are shown in Table 5.

### 4.6 Combined Experiments

As a combined experiment, the pseudo-parallel datasets which produced the best score for the rare word augmentation experiment and dictionary augmentation experiment were combined with the existing parallel-corpus to train the NMT model. Results of this experiment are shown in Table 5.

## 5 Results and Discussion

Our best scores exceeded the baselines of Fadaee et al. (2017) and Peng et al. (2020) which shows that DA benefits by considering both syntactic and semantic constraints. Further random replacements gave insignificant gains. In duplication experiments, the random sample 25K gave an increased BLEU score of +0.23 and +0.41 respectively, however, when the sample size was further increased, the scores were reduced. This means mere duplication was not effective when compared with the DA technique. The experiment was conducted with dictionary terms as well. The results are shown in Table 4. However such random augmentation did not produce any gain but reduced the final BLEU score than the baseline.

### 5.1 Rare Word Augmentation

According to results in Table 4, the comparable results between experiment (Baseline+wordSim) and (Baseline+pos+morph) confirms that for low resource languages, the usage of embeddings is promising in the context of DA task. We further observe that combining the semantic constraints and syntactic constraints leads to improving the results in both directions.

We expected the experiments using pos+morph to be better than the experiment using only pos. As the Sinhala morphological parser did not have the morphological information for some rare words, we observed a reduction in the scores in the Si↦En. However, in the En↦Si direction, experiments using pos+morph improved the scores compared to experiments using pos only.

| Rare word | පරිශීලනය  (pariśīlanaya) |
|---|---|
| Si Sentence | විනිශ්චයකාරවරුන්ගේ පරිශීලනය පිණිස පුස්තකාලය සඳහා 'නීතිය' පිළිබඳ නව ග්‍රන්ථ මිල දී ගන්නා ලදී<br>viniścayakāravarungē **pariśīlanaya** pinisa pustakālaya saňdahā 'nītiya' pilibaňda nava grantha mila dī gannā ladi. |
| En Sentence (Ref.) | New books on "Law" were purchased for the library for the **reference** of the judges. |
| Baseline[train54K] | new law for the library for the library for the library was purchased. |
| Baseline+pos+morph | new law Books were purchased for the Library **reference** to the Judges. |
| Baseline+wordSim | new law Books were purchased on the Library **reference** for the Library reference. |
| Baseline+wordSim+pos | new law Books were purchased for the Library for easy **reference** of the Judges. |

Table 6: Example shows the translation(En) obtained for the Si sentence, from NMT models trained with different augmented sets. These NMT models corresponds to DA techniques considering syntactic constraints (Baseline+pos+morph), semantic constraints (Baseline+wordSim) and both syntactic and semantic constraints (Baseline+wordSim+pos)

With the rare word augmentation, we have obtained the highest BLEU score gain of +0.91 in the Si→En and +0.74 in the En→Si direction.

## 5.2 Dictionary Augmentation

The results obtained for dictionary augmentation in Table 5 show reduced results for Si↦En direction. But this result gradually improved when syntactic and semantic constraints were used. However, our results still exceed the baseline scores of Fadaee et al. (2017) and Peng et al. (2020). From the dictionary augmentation experiments, we obtained a highest gain of +0.71 BLEU in the En→Si direction.

It was observed that the dictionary augmented datasets, more than 12K sentences, always gave a score lesser than the baseline. This was observation was true for both translation directions. Further, the only DA experiment Baseline+wordSim+pos+morph generating 7K sentences produced the highest gain of +0.71 BLEU points. Surprisingly, in rare word DA experiments also, when the synthetic data was more than 8K the scores reduced. Therefore we believe there is a negative effect when the synthetic sentences are high, specifically more than 12K.

In the combined experiments, the parallel augmented sentences were 19K and 7K in the Si→En and En→Si directions respectively. Consistent with the previous behaviour, the BLEU score gain was observed only in En→Si direction as +0.58. However, we need to conduct more experiments to identify the optimal ratio to be maintained between the synthetic sentences and the parallel corpus to achieve the best scores.

Considering the number of rare words and dictionary terms present in the test sets as in Table 2, we observe that even if the OOV terms are present in low counts, the DA technique is still effective enough to improve the overall BLEU score. This is mainly owing to the improvement in the overall translation and its fluency. This is further evident in the example in Table 6.

## 5.3 Qualitative Analysis

The objective of the research was to improve the translations of sequences having OOV terms. To analyze this, we consider a Si sentence, containing a rare word and observed how the translation changes with the DA experiment. As shown in Table 6 the selected sentence has the rare word ie. පරිශීලනය  (reference). We observe that the correct translation of the rare word is generated when using only syntactic or only semantic constraints. However, a more fluent output is generated when both constraints are combined.

Therefore it is evident that DA aids the NMT to improve the translation output for the sequences containing OOV terms.

## 6 Conclusion

This paper presented two data augmentation techniques to generate synthetic sentences in order to address the OOV problem. We further showed that these techniques can be further improved by incorporating syntactic and semantic constraints.

We showed that only using semantic constraints provides results on par with the scores obtained with syntactic constraints. This is promising for low resource languages that have limited linguistic tool support, as semantic constraints can easily be generated with word embeddings learned on monolingual corpora. We also prove that by combining both syntactic constraints and semantic constraints, the scores can be further improved.

However, relying on statistical models for word alignment and for language modeling make the DA sub-optimal. therefore as future work, we will explore multilingual embeddings to improve the word-alignment task and will incorporate a Neural language model instead of a tri-gram statistical language model. Further, we can train a BERT-based model to determine sentence embeddings for Sinhala and explore how those embeddings can be better utilized for the sentence similarity task.

## 7 Acknowledgement

This research was supported by the Accelerating Higher Education Expansion and Development (AHEAD) Operation of the Ministry of Education funded by the World Bank.